\documentclass[journal,final,letterpaper,twocolumn]{IEEEtran}
\usepackage{amsfonts}
\usepackage{amsmath}
\usepackage{array}
\usepackage{epsfig}
\usepackage{subfigure}
\usepackage{amssymb}
\usepackage{threeparttable}
\usepackage{booktabs}
\usepackage{multirow}
\usepackage{color}

\ifCLASSINFOpdf
\fi

\begin{document}
%
\title{Max-Margin based Discriminative Feature Learning}

\author{Changsheng Li,
        Qingshan~Liu,~\IEEEmembership{Senior Member,~IEEE,}
        Weishan~Dong,
        Fan~Wei,
        Xin~Zhang,
        and~Lin~Yang
\thanks{Manuscript received May 21, 2015; revised August 04, 2015; accepted January 10,
2016. This work was supported in part by the National Natural Science Foundation of China under Grant 61532009, 61272223, and in part by the Foundation of Jiangsu Province of China under Grant 15KJA520001.}
\thanks{C. Li, W. Dong, X. Zhang, and L. Yang are with IBM Research-China, Beijing 100094, China.
(email: lcsheng, dongweis, zxin, ylyang$@$cn.ibm.com).}
\thanks{F. Wei is with Department of Mathematics, Stanford University. E-mail: fanwei@stanford.edu.
}
\thanks{Q. Liu is with the B-DAT Laboratory at the school of Information and Control, Nanjing University of Information Science and Technology, Nanjing 210014, China.
(email: qsliu@nuist.edu.cn).}
}

\markboth{}%
{Li \MakeLowercase{\textit{et al.}}: Max-Margin based Discriminative Feature Learning}

\maketitle

\begin{abstract}
In this paper, we propose a new max-margin based discriminative feature learning method. Specifically, we aim at learning a low-dimensional feature representation, so as to maximize the global margin of the data and make the samples from the same class as close as possible. In order to enhance the robustness to noise, we leverage a regularization term to make the transformation matrix sparse in rows. In addition, we further learn and leverage the correlations among multiple categories for assisting in learning discriminative features. The experimental results demonstrate the power of the proposed method against the related state-of-the-art methods.
\end{abstract}


\section{Introduction}
Data classification plays a key role in many practical applications \cite{makinen2008evaluation,mavroforakis2007geometric}.
However, the real-world data, such as image data, often lie in a high-dimensional space, which has high computational cost and might bring down the prediction accuracy of classification models. To cope with this issue, a popular way is to do dimensionality reduction, which is to project the data into a low-dimensional subspace with the least information loss \cite{cai2007spectral}. Generally speaking, dimensionality reduction
can be achieved by either feature selection or feature transformation.
Feature selection aims at selecting the most informative feature subset from the original feature set according to a predefined selection criterion. A number of selection criteria have been proposed in the past several years, such as mutual information \cite{estevez2009normalized},  kernel alignment \cite{shawe2002kernel}, sparsity regularization based measures \cite{nie2010efficient,hou2014joint}, and so on.
The philosophy behind feature transformation is that a combination of the original features may be more helpful for learning. Hence it aims to map the high-dimensional data into a new meaningful low-dimensional space.
Many feature transformation methods have been proposed over the last decades, including manifold learning \cite{tenenbaum2000global,roweis2000nonlinear}, low-rank representation (LRR) \cite{liu2010robust,ni2010robust}, and sparse representation (SR) \cite{huang2010dimensionality,pan2014dimensionality}.
According to the characteristics of the mapping, feature transformation techniques can be further grouped into linear \cite{he22,su17,he23, guoquan} and nonlinear ones \cite{scholkopf1998nonlinear,tenenbaum2000global,roweis2000nonlinear,hou2009stable}.
In this paper, we focus on the linear feature transformation methods due to its simplicity and effectiveness \cite{pechenizkiy2004pca}.

In literature, principal component analysis (PCA) \cite{su17} and linear discriminant analysis (LDA) \cite{duda1999pattern} are two classical linear algorithms, both capturing global Euclidean structure of the data. Since the practical data often lie on or are close to an intrinsically low-dimensional manifold, volumes of approaches focus on how to preserve such structures in recent years. The representative approaches include locality preserving projection (LPP)~\cite{he22}, and neighborhood preserving embedding (NPE)~\cite{he23}, locality sensitive discriminant analysis (LSDA) \cite{caideng}, and stable orthogonal local discriminant embedding (SOLDE)~\cite{guoquan}. Meanwhile, some work integrates global and local information during feature transformation, such as LPMIP \cite{wanghai}, LapLDA \cite{chenjian}.
However, the methods above do not consider how to effectively connect with the classifier in the context of classification. In order to alleviate this limitation, the maximum margin projection (MMP) algorithm \cite{wang2011unsupervised} takes advantage of a binary support vector machine (SVM) \cite{vapnik2000nature} classifier to obtain some hyperplanes that separate data points in different clusters with the maximum margin. The random projection algorithms \cite{majumdar2010robust,paul2013random} aim to find some Gaussian random projection matrices to preserve the pairwise distances between data points in the projected subspace, which can be effectively combined with some classifiers, such as SVM. {Varshney and Willsky \cite{varshney2011linear} propose a framework to simultaneously learn a linear dimensionality reduction projection matrix and a margin-based classifier defined in the reduced space. The main idea of the maximum margin projection pursuit (MMPP) algorithm is to integrate optimal data embedding and SVM classification in a single framework in both bi-class and multi-class classification \cite{nikitidis2014maximum}. In addition, local discriminant gaussian (LDG)~\cite{parrish} is a feature transformation method which exploits a smooth approximation of the leave-one-out cross validation error of a classifier.
However, these methods do not intend to preserve the intrinsic manifold structure of the data and removing noisy features before feature transformation. In addition, there are often correlations among multiple classes in practical scenarios. Previous works do not study how to leverage such correlations with feature transformation. Gu et al. \cite{gu2011joint} propose a framework for joint subspace learning and feature selection, called FSSL, which can alleviate the effect of the noisy features for feature transformation. However, they do not focus on how to effectively combine the classifier in the scenario of classification.

In this paper, we propose a new \underline{M}ax-\underline{M}argin based feature transformation method to \underline{L}earn \underline{D}iscriminative \underline{F}eatures for classification, called MMLDF. In the learned low-dimensional feature space, our method aims at maximizing the global classification margin; in the meantime, the distances of samples from the same class are minimized.
Additionally, in many real-world applications, there are often correlations among multiple classes, and capturing such correlations are helpful for learning discriminative features and designing classifiers \cite{evgeniou2007multi}. In light of this point, we add a regularization term to capture the correlations among multiple categories for feature learning. Extensive experiments are conducted on eight publicly available datasets, and the experimental results demonstrate the effectiveness of the proposed method against the state-of-the-art methods.

The rest of the paper is organized as follows. Section II gives the details of the proposed method. The experimental results are reported and analyzed in Section III. Section IV concludes the paper.
\section{Proposed Method}
Let $\mathcal{X}=\{(\mathbf{x}_i,y_i)\}_{i=1}^n$ denote a training data set, where $\mathbf{x}_i\in \mathbb{R}^d$ is the $i$-th  data point and $y_i\in \{1, \ldots, \mathcal{K}\}$ represents the corresponding class label. Our goal is to obtain a projection matrix $\mathbf{P}\in \mathbb{R}^{d\times r}$ that maps the $d$-dimensional input vector to an $r$-dimensional vector ($r< d$) by $\mathbf{z}_i=\mathbf{P}^T\mathbf{x}_i$. For the transformation matrix $\mathbf{P}$, let $\mathbf{p}_i$ represent its $i$-th row, and $P_{ij}$ denote its $(i,j)$-th entry. As usual, we use $tr(\mathbf{P})$ to denote the trace of $\mathbf{P}$, and $\| \mathbf{P}\|_{2,1}$ denotes $l_{2,1}$ norm of $\mathbf{P}$ defined as:
\vspace{0.1in}
\begin{align}\label{p21}
 \| \mathbf{P}\|_{2,1}=\sum_{i=1}^d \| \mathbf{p}_i\|_{2}=\sum_{i=1}^d\sqrt{\sum_{j=1}^rP_{ij}^2}
\end{align}

\subsection{Binary Classification: $\mathcal{K}=2$}
Margin maximization has been demonstrated to be a good principle applied by various learning methods~\cite{weinberger,Burgesc}. Among these methods, support vector machine (SVM) is a popular one that maximizes the global margin of data points of different classes, but it only focuses on how to learn a max-margin based classifier.  In this paper, we aim at learning a low-dimensional feature space to further improve the performance of max-margin based classification, and we present a unified framework to integrate feature learning and classification together. To do this, we propose the following objective function:
\begin{align}\label{equ:objg}
\min J({\mathbf{P},\mathbf{w},b})
\!&=\!\frac{1}{2}\|\mathbf{w}\|^2\!+\!C \sum_{i=1}^n l(\mathbf{w},b,\mathbf{P};\mathbf{x}_i,y_i)\! \nonumber\\
&\!+\!\eta\sum_{i,j=1}^n\!\|\mathbf{P}^T\mathbf{x}_i\!-\!\mathbf{P}^T\mathbf{x}_j\|^2A_{ij}\!+\!\!\lambda\|\mathbf{P}\|_{2,1}
\end{align}
where $C$, $\eta$ and $\lambda$ are three trade-off parameters, in order to balance the contribution of each term to learn discriminative features. Through these parameters, our model can be flexible to different scenarios.  $\mathbf{w}$ is the learned weight vector, as in SVM. $\mathbf{P}$ is the learned transformation matrix which projects the original data into low-dimensional feature space. $\mathbf{P}^T\mathbf{x}_i$ is the new low-dimensional representation of $\mathbf{x}_i$. $A_{ij}$ denotes the edge weight of the within-class graph. Here, either Heat kernel or Simple-minded can be used for weighting the edges. For similarity, we choose Simple-minded in our experiment, which is expressed as:
\vspace{0.07in}
\begin{align}
A_{ij}=
\left\{
\begin{array}{cc}
1 & \mbox{if}\; y_i=y_j \; \\
0 & \mbox{otherwise}
\end{array}
\right.
\end{align}
\vspace{0.02in}

$l(\bullet)$ is a hinge loss function. The standard hinge loss function in SVM is not differentiable everywhere.
In order to take advantage of the gradient-based optimization method to solve the proposed objective function, we adopt a similar but differentiable quadratic hinge loss:
\vspace{0.07in}
\begin{align}
l(\mathbf{w},b,\mathbf{P};\mathbf{x}_i,y_i)=[\min(0,y_i\!\cdot\!(\mathbf{w}^T\!\cdot\! (\mathbf{P}^T\mathbf{x}_i)\!+\!b)\!-\!1)]^2
\end{align}
\vspace{0.02in}

 Minimizing the first two terms of (\ref{equ:objg}) means finding a low-dimensional subspace, in which the margin of different classes can be maximized, and the classification loss can be minimized. Minimizing the third term makes the scatter of the data in the same class as small as possible in the subspace. The last term is a regularization term to make the transformation matrix $\mathbf{P}$ sparse in rows, so that the learned subspace is robust to noise, and the model complexity is lowered too.

 We take the LSVT Voice Rehabilitation dataset \cite{Tsanas2014}, a bi-class classification dataset, as an example to illustrate the effectiveness of the $l_{2,1}$ norm constraint on $\mathbf{P}$ in the objective function (\ref{equ:objg}). Fig. \ref{visuall21} (a) and (b) show the visualizations of $\mathbf{P}$ without $l_{2,1}$ norm and with $l_{2,1}$ norm in (\ref{equ:objg}), respectively. We can see that many rows in $\mathbf{P}$ become sparse by adding the $l_{2,1}$ norm constraint, which can eliminate noisy features in the process of feature transformation, and can reduce the model complexity. In the later experiment, we will further demonstrate that our method is robust to noise by introducing $l_{2,1}$ norm.
\begin{figure}
  \center
\subfigure[Without $l_{2,1}$ norm constraint]{\includegraphics[width=0.49\linewidth]{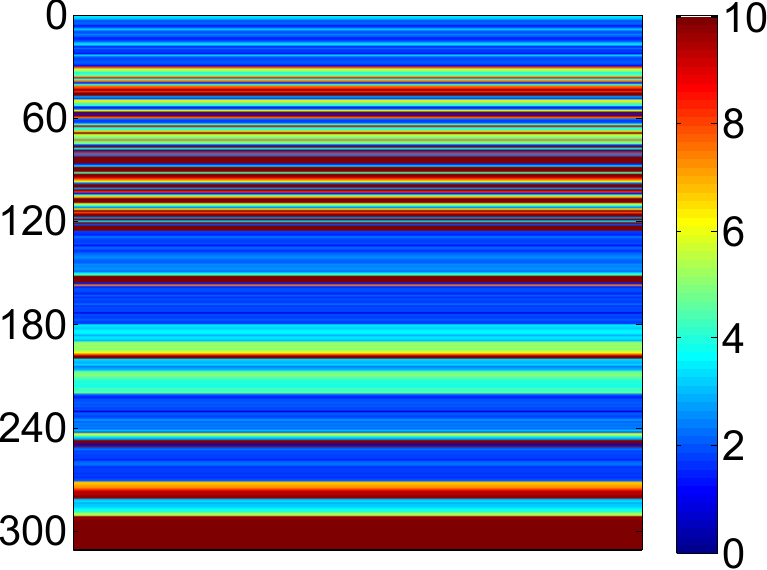}}
\subfigure[With $l_{2,1}$ norm constraint]{\includegraphics[width=0.49\linewidth]{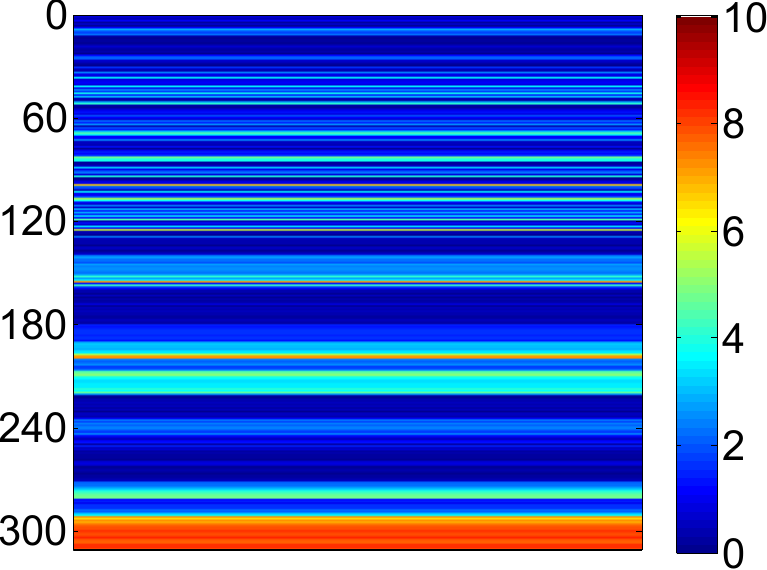}}
\caption{The visualization of the learned $\mathbf{P}\ (r=10)$ on the LSVT Voice Rehabilitation dataset. Each row is the $l_2$ norm value of each row of $\mathbf{P}$. Dark blue denotes that the values are close to zero.}
\label{visuall21}
\end{figure}

\subsection{Multi-class Classification: $\mathcal{K} > 2$}
In the case of multi-class classification, we can extend (\ref{equ:objg}) to the following objective function:
\begin{align}\label{total_obj2}
\min&\  J_{mc}({\mathbf{W},\mathbf{b},\mathbf{P}}) \nonumber\\
=&\frac{1}{2}\sum_{m=1}^{\mathcal{K}}\|\mathbf{w}_m\|^2+C \sum_{i=1}^n \sum_{m\neq y_i} l(\mathbf{w}_m, \mathbf{w}_{y_i}, b_{y_i}, b_m,\mathbf{P};\mathbf{x}_i, y_i) \nonumber\\
&+\eta\sum\nolimits_{i,j=1}^n\|\mathbf{P}^T\mathbf{x}_i-\mathbf{P}^T\mathbf{x}_j\|^2 A_{ij}+\lambda\|\mathbf{P}\|_{2,1}
\end{align}
where $\mathbf{W}=[\mathbf{w}_1, \ldots, \mathbf{w}_{\mathcal{K}}]$ is the set of the learned weight vectors. $l(\mathbf{w}_m, \mathbf{w}_{y_i}, b_{y_i}, b_m,\mathbf{P};\mathbf{x}_i, y_i)$ measures the loss when the sample $\mathbf{x}_i$ is wrongly classified into the $m$-th $( m\neq y_i)$ class. Similarly, the loss function $l$ is revised as:
\begin{align}
&l(\mathbf{w}_m, \mathbf{w}_{y_i}, b_{y_i}, b_m,\mathbf{P};\mathbf{x}_i, y_i)\nonumber\\
&=[\min(0,(\mathbf{w}_{y_i}^T\mathbf{P}^T\mathbf{x}_i+b_{y_i}-\mathbf{w}_m^T\mathbf{P}^T\mathbf{x}_i-b_m-2)]^2
\end{align}

In real-world applications, one classification task is often correlated with other classification tasks, and mining the correlations among multiple categories can be good for feature learning \cite{evgeniou2007multi}. Our experiments also demonstrate this point (See the details in \ref{comp_role}). Thus, we add a regularization term into (\ref{total_obj2}) to capture the correlations among multiple categories, and the new objective function becomes:
\begin{align}\label{total_obj3}
&\min \  J_{mc}({\mathbf{W},\mathbf{b},\mathbf{P},\Gamma}) \nonumber\\
&=\frac{1}{2}\sum_{m=1}^{\mathcal{K}}\|\mathbf{w}_m\|^2 \!+\!C \!\sum_{i=1}^n\! \!\sum_{m\neq y_i}\! l(\mathbf{w}_m,\mathbf{w}_{y_i},b_{y_i},b_m,\mathbf{P};\mathbf{x}_i,y_i)   \nonumber\\
&+\eta \!\sum_{i,j=1}^n \! \|\mathbf{P}^T\mathbf{x}_i\!-\!\mathbf{P}^T\mathbf{x}_j\|^2A_{ij}\!+\!\lambda\|\mathbf{P}\|_{2,1}\!+\!\rho \|\mathbf{W}^T\|_{\Gamma}^2
\end{align}
where $\rho$ is a trade-off parameter. $\|\mathbf{W}^T\|_{\Gamma}^2=tr(\mathbf{W}\Gamma\mathbf{W}^T)$ is the Mahalanobis norm of the matrix $\mathbf{W}^T$. $\Gamma$ plays the role of the inverse covariance matrix that encodes the correlations among the weight vectors $\mathbf{w}_i$ \cite{zhang2012convex}. The inverse matrix of $\Gamma$ is constrained to be positive definite and unit trace, in order to obtain a valid solution. The effect of the last term in (\ref{total_obj3}) is to  penalize the complexity of $\mathbf{W}$ relying on the Mahalanobis norm, as well as to learn the inverse covariance matrix $\Gamma$ simultaneously.

We use the Urban Land Cover dataset \cite{johnson2013classifying}, a multi-class classification dataset, to visualize the correlation coefficient matrix of the weight vectors, which can be obtained based on the learned $\Gamma$. The result is shown in Fig. \ref{correlation_matrix}. We can see that there are indeed correlations among the multiple categories (e.g. the second and the third weight vectors).
\begin{figure}
\center
\includegraphics[width=0.45\linewidth]{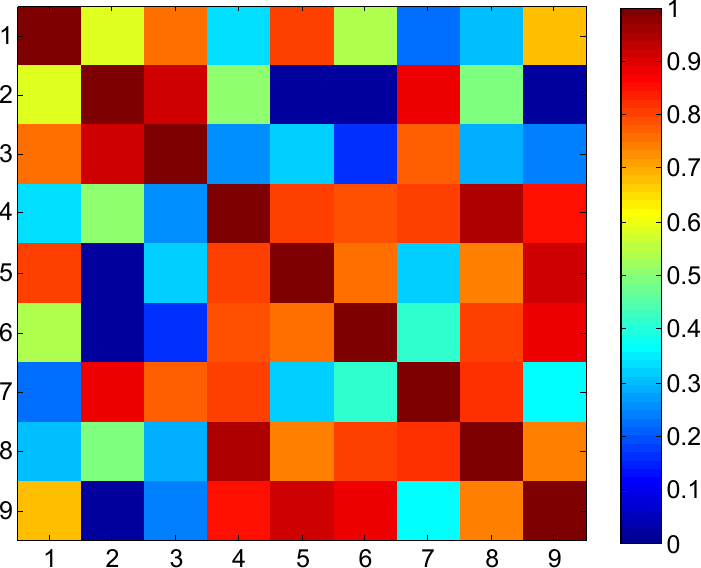}\\
\caption{ The visualization of the learned correlation coefficient matrix on the Urban Land Cover dataset. Both x-axis  and  y-axis  represent  weight vectors. Dark red denotes that the correlation coefficients are close to 1.}\label{correlation_matrix}
\end{figure}
\subsection{Optimization Procedure}
We first introduce how to solve the optimization problem in the binary classification case.
The objective function (\ref{equ:objg}) is not convex with respect to the variables $\mathbf{w}$, $b$, and $\mathbf{P}$ simultaneously. Therefore it is unrealistic to expect an algorithm to easily find the global minimum of $J$. Therefore, we adopt an alternating optimization strategy to find the local minimum. Under this scheme, we update $\mathbf{P}$, $\mathbf{w}$, and $b$ in an alternating manner.

{Update} $\mathbf{P}$, with fixed $\mathbf{w}$ and $b$:  When $\mathbf{w}$ and $b$ are fixed, (\ref{equ:objg}) becomes a convex problem, so $\mathbf{P}$ can be obtained by minimizing the following objective function:
\begin{align}
J_1(\mathbf{P})=&C \sum_{i=1}^n [\min(0,y_i\cdot(\mathbf{w}^T\cdot (\mathbf{P}^T\mathbf{x}_i)\!+\!b)\!-\!1)]^2\nonumber\\
&\!+\!\eta\sum_{i,j=1}^n\|\mathbf{P}^T\mathbf{x}_i\!-\!\mathbf{P}^T\mathbf{x}_j\|^2A_{ij}+\lambda\|\mathbf{P}\|_{2,1}\nonumber
\end{align}

Taking the derivative of $J_1$ with respect to $\mathbf{P}$, we can obtain:
\begin{align}\label{derives1}
\frac{\partial J_1(\mathbf{P})}{\partial \mathbf{P}}=&2C\!\sum_{(\mathbf{x}_i,y_i)\in \Theta}\!(\mathbf{x}_i\mathbf{x}_i^T\mathbf{P}\mathbf{w}\mathbf{w}^T\!-\!(y_i\!-\!b)\mathbf{x}_i\mathbf{w}^T)\nonumber \\
&+2\eta\mathbf{X}\mathbf{L}\mathbf{X}^T\mathbf{P}+2\lambda\mathbf{D}_P\mathbf{P}
\end{align}
where $\mathbf{X}=[\mathbf{x}_1,\ldots,\mathbf{x}_n]$. $\mathbf{L}=\mathbf{D}_A-\mathbf{A}$ is the Laplacian matrix, where $\mathbf{D}_A$ is a diagonal matrix with its entries being row sum of $\mathbf{A}$, $D_A(i,i)=\Sigma_jA_{ij}$. $\mathbf{D}_P$ is a diagonal matrix with the $i$-th diagonal element $\mathbf{D}_P(i,i)=\frac{1}{2\|\mathbf{p_i}\|_2}$. $\Theta$ denotes the set of $(\mathbf{x}_i,y_i)$ satisfying the following condition: $y_i(\mathbf{w}^T(\mathbf{P}^T\mathbf{x}_i)+b)-1\leq0$\footnote{When $\Theta=\phi,$ we define $\sum\limits_{(\mathbf{x}_i,y_i)\in \Theta}\!(\mathbf{x}_i\mathbf{x}_i^T\mathbf{P}\mathbf{w}\mathbf{w}^T\!-\!(y_i\!-\!b)\mathbf{x}_i\mathbf{w}^T)=\mathbf{0}$.}.

Setting the derivative in (\ref{derives1}) to zero, these is no closed-from solution of $\mathbf{P}$. Therefore, we adopt a gradient-based method to derive the optimal $\mathbf{P}$. Here we choose the limited-memory BFGS (L-BFGS) algorithm for its efficiency~\cite{qianming,liudeng}, which is summarized in Algorithm 1.
The core idea in Algorithm 1 is to estimate the inverse Hessian matrix $\mathbf{H}$ only using the latest $m$ updates of the position $\mathbf{p}_k$ and gradient $\mathbf{g}_k$ to lower memory complexity, where $m$ is usually small.

\begin{table}[htb]
\small
\begin{center}
\label{tableMAEs}
\begin{tabular}{l}
\hline
\textbf{Algorithm 1} \   L-BFGS \\
\hline
\textbf{Input:} Randomly initialize $\mathbf{P}_0\in \mathbb{R}^{d\times r}$, and reshape $\mathbf{P}_0$ into \\
 \ \ \ \ \ \ \ \ \  a vector $\mathbf{p}_0$; Randomly initialize a symmetric and positive\\
  \ \ \ \ \ \ \ \ \   definite matrix $\mathbf{H}_0$;   Set the integer $m=25$.\\
 \ \ \ \ \ \ \ \ \ $k\leftarrow0$\\
\textbf{Output:} $\mathbf{P}$\\
\ \textbf{Repeat} \\
\ \ \ \ \ \ Compute the gradient $\mathbf{G}_k=\frac{\partial J_1(\mathbf{P})}{\partial \mathbf{P}}|_{\mathbf{P}_k}$, and reshape $\mathbf{G}_k$ \\
\ \ \ \ \ \ \ into a vector $\mathbf{g}_k$.\\
 \ \ \ \ \ \ Compute $d_k\leftarrow -\mathbf{H}_k\cdot \mathbf{g}_k$ using a two-loop recursion; \\
 \ \ \ \ \ \ Compute $\mathbf{p}_{k+1}\leftarrow \mathbf{p}_k+\alpha_kd_k$, where $\alpha_k$ satisfies \\
 \ \ \ \ \ \ \ the Wolfe conditions;\\
 \ \ \ \ \ \ $\mathbf{s}_k\leftarrow \mathbf{p}_{k+1}-\mathbf{p}_k,\mathbf{y}_k\leftarrow\mathbf{g}_{k+1}- \mathbf{g}_k$;\\
 \ \ \ \ \ \ $\widehat{m}\leftarrow \min \{k,m-1\}$; \\
 \ \ \ \ \ \ Update Hessian matrix $\mathbf{H}_k$ using the pairs $\{\mathbf{y}_j,\mathbf{s}_j\}_{j=k-\widehat{m}}^k$;\\
 \ \ \ \ \ \ Reshape $\mathbf{p}_{k+1}$ into a matrix $\mathbf{P}_{k+1}$;\\
 \ \ \ \ \ \ $k\leftarrow k+1$;\\
 \ \textbf{Until} Convergence criterion satisfied; \\
 \ $\mathbf{P}\leftarrow \mathbf{P}_k$.\\
\hline
\end{tabular}
\end{center}
\end{table}


{Update} $\mathbf{w}$, with fixed $b$ and $\mathbf{P}$: When $b$ and $\mathbf{P}$ are fixed, (\ref{equ:objg}) is convex in terms of $\mathbf{w}$, so we minimize the following objective function to obtain the optimal $\mathbf{w}$:
\begin{align}
 J_2(\mathbf{w})\!=\!\frac{1}{2}\|\mathbf{w}\|^2\!+\! C \sum_{i=1}^n [\min(0,y_i\cdot(\mathbf{w}^T\cdot (\mathbf{P}^T\mathbf{x}_i)\!+\!b)\!-\!1)]^2\nonumber
\end{align}

Taking the derivative of $J_2(\mathbf{w})$ with respect to $\mathbf{w}$, and setting it to zero, we obtain:
\begin{align}\label{solu_w}
\mathbf{w}
=\left\{
\begin{array}{cc}
(\mathbf{I}+2C\sum\limits_{(\mathbf{x}_i,y_i)\in \Theta}\mathbf{P}^T\mathbf{x}_i\mathbf{x}_i^T\mathbf{P}\mathbf{w})^{-1} & \\
\times(2C\mathop\sum\limits_{(\mathbf{x}_i,y_i)\in \Theta}(y_i-b)\mathbf{P}^T\mathbf{x}_i) & \mbox{if}\; (\mathbf{x}_i,y_i)\in \Theta\\
0 & \mbox{otherwise}
\end{array}
\right.
\end{align}

{Update} $b$, with fixed $\mathbf{w}$ and $\mathbf{P}$: When $\mathbf{w}$ and $\mathbf{P}$ are fixed, (\ref{equ:objg}) is convex in terms of $b$, so $b$ can be acquired by optimizing the following objective:
\begin{align}
\min J_3(b)=\sum\nolimits_{i=1}^nl(b;\mathbf{w},\mathbf{P}^T\mathbf{x}_i,y_i)\nonumber
\end{align}

Taking the derivative of $J(b)$ with $b$, and setting it to zero, we obtain:
\begin{align}\label{solut_b}
b=\left\{
\begin{array}{cc}
\frac{\sum\limits_{(\mathbf{x}_i,y_i)\in\Theta} (y_i-\mathbf{w}^T\mathbf{P}^T\mathbf{x}_i)}{\mid\Theta\mid} & \mbox{if}\;  \Theta \neq\phi\\
b & \mbox{otherwise}
\end{array}
\right.
\end{align}
where $|\Theta|$ denotes the size of the set $\Theta$.

The procedure of the proposed algorithm can be summarized in Algorithm 2. We randomly initialize the parameters, and adopt the second updating rule in \cite{hou2014discriminative} for deriving the local optimal solution. Experimental results verify that MMLDF converges quickly and obtains promising local minimums.

In the multi-classification case, we still adopt the alternating optimization strategy to find the local minimum. We take advantage of the L-BFGS to update $\mathbf{P}$ and $\mathbf{w}_i$. $b$ still has a closed-form solution derived by using the same optimization method with the binary classification case.
The rule for updating the variable $\Gamma$ in (\ref{total_obj3}) is as follows:
\begin{align}\label{solu_theta}
\Gamma=\frac{(\mathbf{W}^T\mathbf{W})^{1/2}}{tr((\mathbf{W}^T\mathbf{W})^{1/2})}
\end{align}

Details of the proof on (\ref{solu_theta}) can be found in \cite{zhang2012convex}.

\subsection{Discussion}
Varshney and Willsky \cite{varshney2011linear} propose a linear dimensionality reduction method, which represents the learned mappings by matrices on the Stiefel manifold and on margin-based classifiers. Nikitidis et al. \cite{nikitidis2014maximum} present the maximum margin projection pursuit (MMPP) algorithm, which learns the optimal data embedding and the SVM classifier simultaneously. However, our approach constructs the graph Laplacian through local learning to capture the intrinsic structure of the data, and incorporates $l_{2,1}$-norm minimization into our framework to alleviate the effects of noisy features. In addition, our method can mine the correlations among multiple categories, which is beneficial to both feature transformation and classifier training. When $\eta\rightarrow0, \lambda\rightarrow0, \rho\rightarrow0$ in (\ref{total_obj3}), our objective function has similar effects to those of \cite{varshney2011linear} and \cite{nikitidis2014maximum}.

Gu et al. \cite{gu2011joint} propose a framework for joint subspace learning and feature selection, where subspace learning is reformulated as solving a linear system equation, and feature selection is achieved by utilizing $l_{2,1}$-norm on the projection matrix. However, they do not explore how to effectively combine the classifier for classification. When $C\rightarrow0, \rho\rightarrow0$, and adding a constraint on the projection matrix in (\ref{total_obj3}), the formulation of our method is reduced to that of \cite{gu2011joint}.
}

Xu et al. \cite{xu2010discriminative} propose a semi-supervised feature selection method called FS-Manifold, where the feature selection process is embedded with a manifold regularized SVM classifier. Different from FS-Manifold, our method aims to incorporate the feature transformation process into the SVM framework.

In addition, MMLDF is easily extended into the kernel version. We take binary classification as an example, and handle the multi-classification case in the same way. According to \cite{Burgesc}, we know the original max-margin objective function of SVM can be transformed into its dual version as:
\begin{align}\label{svmdual}
&\max \theta(\alpha)=\sum_i\alpha_i-\frac{1}{2}\sum_i\sum_j\alpha_i\alpha_jy_iy_j\mathbf{x}_i^T\mathbf{x}_j\\
&s.t.\ \ \sum_i\alpha_iy_i=0, \alpha_i\geq0 \nonumber
\end{align}

Based on (\ref{svmdual}), we can incorporate the feature transformation process into the kernel SVM framework as:
\begin{align}\label{svmduald}
\max \theta (\alpha,\mathbf{P})=&\sum_i\alpha_i-\frac{1}{2}\sum_i\sum_j\alpha_i\alpha_jy_iy_jk(\mathbf{P}^T\mathbf{x}_i,\mathbf{P}^T\mathbf{x}_j)\nonumber\\
&-\eta\sum_{i,j=1}^n\!\|\mathbf{P}^T\mathbf{x}_i\!-\!\mathbf{P}^T\mathbf{x}_j\|^2A_{ij}\!-\!\!\lambda\|\mathbf{P}\|_{2,1}\\
s.t.\ \ \sum_i\alpha_iy_i&=0, \alpha_i\geq0 \nonumber
\end{align}
where $k(\mathbf{P}^T\mathbf{x}_i,\mathbf{P}^T\mathbf{x}_j)$ is the kernel function, such as the radial basis function (RBF),  polynomial kernel, and so on.
\subsection{Time Complexity Analysis}
The time complexity of Algorithm 2 consists of three parts: initialization on line 1, Laplacian matrix construction on line 2, and the iterative update of the three variables on lines 3-8. The complexities of the first two parts can be ignored compared to the third part. In the third part, we need to update $\mathbf{w}$, $b$, and $\mathbf{P}$, respectively. For updating $\mathbf{w}$, the worst complexity is $O(ndr+r^3)$. Updating $b$ costs $O(ndr)$. For updating $\mathbf{P}$, it needs $O(t_1*mdc)$ for the two-loop recursion scheme, where $t_1$ denotes the total number of iterations. According to (\ref{derives1}), the worst case of computing partial gradient w.r.t. $\mathbf{P}$ is $O(t_2*(nd^2+n^2d))$, where $t_2$ is the total number of computing the gradient. The complexity of evaluating the objective function values is $O(t_3*(rn^2+ndr+r^2d))$, where $t_3$ denotes the total number of evaluations. Therefore, the total time complexity of MMLDF is of order $O(t*(ndr+r^3)+t_1*mdc+t_2*(nd^2+n^2d)+t_3*(rn^2+ndr+r^2d))$, where $t$ is the total number of iterations in Algorithm 2. Since $t<t_3$ and $r<d$, the complexities of the parts updating $\mathbf{w}$ and $b$ can be ignored, compared to that of updating $\mathbf{P}$, i.e., the time complexity of MMLDF is dominated by updating $\mathbf{P}$.
\begin{table}
\small
\begin{center}
\label{algorithm}
\begin{tabular}{l}
\hline
\textbf{Algorithm 2} \   Max-Margin based Discriminative Feature Learning \\
\hline
\textbf{Input:} Training dataset $\mathcal{X}=\{\mathbf{x}_i,y_i\}_{i=1}^n$;\\
 \ \ \ \ \ \ \ \ \ The parameters: $C, \lambda, \eta, \rho$; \\
 \ \ \ \ \ \ \ \ \ Reduced dimension $r$\\
 \textbf{Method}\\
1. \ \ \ Initialize iteration step $t=0$; Randomly initialize \\
\ \ \ \ \ \ \ $\mathbf{w}^t, b^t$, $\mathbf{P}^t$;\\
2. \ \ \ Construct Laplacian matrix $\mathbf{L}$ ;\\
3. \ \ \ \textbf{Repeat} \\
4. \ \ \ \ \ \ Fixing $\mathbf{P}^t$ and $b^t$, update $\mathbf{w}^{t\!+\!1}$ by Eq. (\ref{solu_w});\\
5. \ \ \ \ \ \ Fixing $\mathbf{P}^t$ and $\mathbf{w}^{t\!+\!1}$, update $b^{t+1}$ by  Eq. (\ref{solut_b});\\
6. \ \ \ \ \ \ Fixing $\mathbf{w}^{t\!+\!1}$ and $b^{t\!+\!1}$, update $\mathbf{P}^{\!t+\!1}$ by Algorithm 1;\\
7. \ \ \ \ \ \ $t=t\!+\!1$;\\
8. \ \ \ \textbf{Until} Convergence criterion satisfied. \\
\textbf{Output:} Transformation matrix $\mathbf{P}\in \mathbb{R}^{d\times r}$\\
\hline
\end{tabular}
\end{center}
\end{table}
\subsection{Evaluation of Convergence Rate}
\begin{figure}
\centering
\subfigure[LSVT Voice Rehabilitation]{\includegraphics[width=0.48\linewidth]{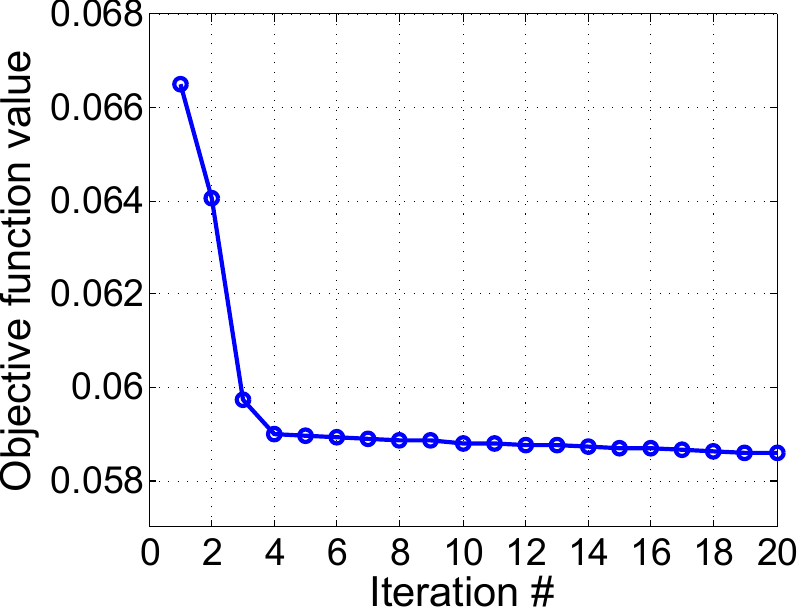}}
\subfigure[Urban Land Cover]{\includegraphics[width=0.48\linewidth]{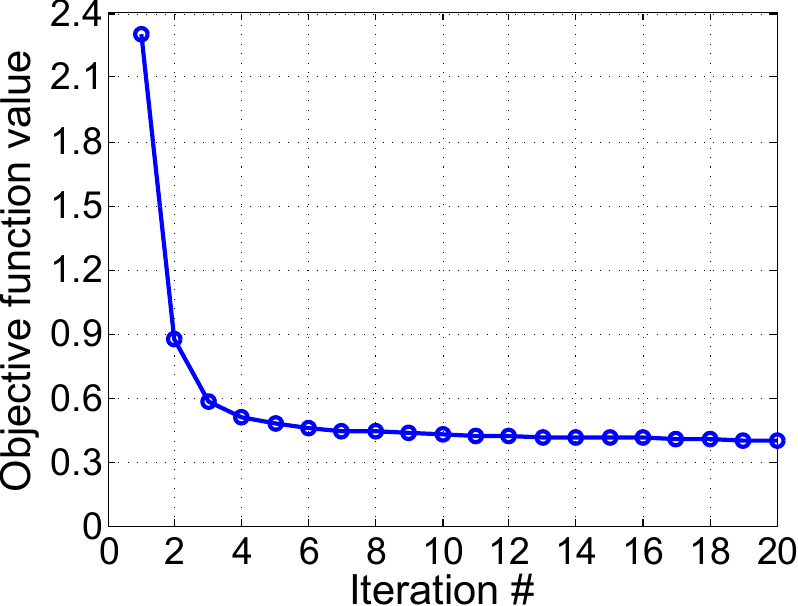}}
\caption{Empirical study on the convergence of MMLDF on the LSVT Voice Rehabilitation dataset and the Urban Land Cover dataset.}
\label{fig:convergence}
\end{figure}

Although the convergence of the MMLDF algorithm cannot be proved theoretically, we find that it converges asymptotically in our experiments.
Fig. \ref{fig:convergence} shows the convergence curves of MMLDF on the LSVT Voice Rehabilitation dataset and the Urban Land Cover dataset, respectively.  As in Fig. \ref{fig:convergence}, we can see MMLDF has a good convergence rate. It will converge after only about 6 iterations.

\begin{table}
\footnotesize
\begin{center}
\caption{Summary of experimental datasets. $\#$Size, $\#$Train, $\#$Test, $\#$Dim, and $\#$Cat denote the number of samples, the number of training, the number of testing, the number of features, and the number of categories, respectively.}
\begin{tabular}{|c|c|c|c|c|c|}
\hline
Dataset & $\#$Size & $\#$Train & $\#$Test &$\#$Dim & $\#$ Cat \\
\hline
Urban Land Cover  & 168  & 42 &126 & 148 & 9 \\
\hline
CNAE-9  & 1,080  & 100 &980 & 857 & 9 \\
\hline
DNA & 2,000 &100&1,900& 180 & 3 \\
\hline
Glioma & 50 &  10  &40& 4,434 & 4 \\
\hline
LSVT Voice Rehab. & 126 & 10 & 116 &309 & 2 \\
\hline
Epsilon & 5,000 & 1,000 &4,000 & 2,000 & 2 \\
\hline
Yale & 165 & 30 & 135 & 512 & 15 \\ \hline
15 scene & 4,485 & 500 &3,985 & 512 & 15 \\ \hline
\end{tabular}
\end{center}
\end{table}

\begin{table*}
\small
\begin{center}
\caption{Classification accuracy(mean+deviation ($\%$)) of different algorithms on all the eight datasets}
\begin{tabular}{|c|c|c|c|c|c|c|c|c|}
\hline
\multirow{2}{*}{ Datasets} & \multicolumn{8}{|c|}{Method}\\
\cline{2-9}
&MMPP&FSSL &LDG &LPMIP &LSDA & SOLDE & LapLDA & MMLDF \\
\hline
Urban Land Cover     &70.6$\pm$3.9 &  49.0$\pm$6.2  & 57.9$\pm$5.5  & 58.8$\pm$5.8  & 55.2$\pm$6.8  & 67.2$\pm$5.3  & 57.9$\pm$4.8 &  \textbf{75.2}$\pm$4.0\\
  & (80) & (40) & (10) & (20) &  (10) & (60)& (9)& (70) \\
\hline
   CNAE-9            &74.4$\pm$2.9&  77.6$\pm$3.6  & 72.9$\pm$2.8  & 79.8$\pm$2.7  & 76.4$\pm$6.9  & 74.2$\pm$3.2  & 78.2$\pm$4.1 &  \textbf{83.6}$\pm$1.6\\
& (70) & (10) & (10) & (80) &  (70) & (40)& (9)& (20) \\
\hline
   DNA               &76.0$\pm$2.5&  72.4$\pm$1.6  & 82.0$\pm$2.8  & 79.4$\pm$2.9  & 72.8$\pm$1.9  & 81.6$\pm$2.2  & 74.9$\pm$1.8 &  \textbf{83.6}$\pm$1.6\\
& (80) & (10) & (10) & (50) &  (10) & (60)& (3)& (10) \\
\hline
   Glioma            &50.5$\pm$6.3&  28.0$\pm$10.1 & 40.8$\pm$8.2  & 46.3$\pm$7.9  & 47.3$\pm$6.0  & 51.5$\pm$6.9  & 49.3$\pm$5.8 &  \textbf{54.0}$\pm$7.3\\
& (30) & (10) & (10) & (10) &  (10) & (20)& (4)& (40) \\
 \hline
   LSVT Voice Rehab. &72.7$\pm$7.5&  71.3$\pm$6.4  & 63.3$\pm$8.3  & 68.7$\pm$6.3  & 68.7$\pm$8.0  & 71.6$\pm$6.5  & 72.4$\pm$7.6 &  \textbf{77.2}$\pm$7.2\\
 & (100) & (40) & (10) & (20) &  (10) & (20)& (2)& (20) \\
  \hline
   Epsilon           &70.6$\pm$1.3&  67.8$\pm$0.8  & 75.7$\pm$0.8  & 74.1$\pm$1.0  & 67.9$\pm$0.8  & 74.6$\pm$0.4  & 70.4$\pm$0.3 &  \textbf{78.2}$\pm$0.6\\
  & (10) & (10) & (100) & (100) &  (10) & (100)& (2)& (10) \\
   \hline
   Yale         &55.0$\pm$4.9&  21.6$\pm$6.5  & 58.8$\pm$6.8  & 60.1$\pm$6.1  & 62.2$\pm$6.4  & 54.3$\pm$4.2  & 60.7$\pm$6.3 &  \textbf{65.5}$\pm$5.1\\
   & (60) & (30) & (20) & (30) &  (20) & (20)& (15)& (40) \\
   \hline
   15 scene          &35.0$\pm$3.6&  17.9$\pm$0.9  & 59.2$\pm$2.1  & 59.0$\pm$2.1  & 8.5$\pm$0.9  & 59.4$\pm$2.0  & 15.7$\pm$1.6 &  \textbf{64.2}$\pm$0.9\\
  & (100) & (20) & (70) & (90) &  (10) & (100)& (15)& (50) \\
  \hline
\end{tabular}
\end{center}
\end{table*}
\section{Experiments}
\subsection{Datasets and Experimental Settings}
We evaluate the performance of MMLDF on eight real-world datasets, including one aerial image dataset Urban Land Cover \cite{johnson2013classifying}, two biomedical area datasets DNA and Glioma, one large scale learning competition dataset Epsilon, one speech signal processing area dataset LSVT Voice Rehabilitation \cite{Tsanas2014}, one business area dataset CNAE-9, one face recognition dataset Yale Face, and one scene classification dataset 15 scene. The datasets DNA and Epsilon are downloaded from LIBSVM official web page\footnote {http://www.csie.ntu.edu.tw/~cjlin/libsvmtools/datasets/}, and the dataset CNAE-9 is downloaded from UCI Machine Learning Repository\footnote{http://archive.ics.uci.edu/ml/datasets/CNAE-9}.  Datasets from different areas serve as a good test bed
for a comprehensive evaluation. Table 1 summarizes the details of the datasets used in the experiments.

To verify the effectiveness of MMLDF, we compare it with the following seven related linear feature transformation methods:
\begin{itemize}
\item
 SOLDE: Stable Orthogonal Local Discriminant Embedding~\cite{guoquan} reduces the dimensions by considering both the diversity and similarity.
 \item
 LDG: Local Discriminant Gaussian~\cite{parrish} exploits a smooth approximation of the leave-one-out cross validation error of a quadratic discriminant analysis classifier\footnote{The MATLAB code for LDG was obtained from the authors of \cite{parrish}}.
 \item
LPMIP: Locality-Preserved Maximum Information Projection~\cite{wanghai} aims to preserve the local structure while maximizing the global information simultaneously.
 \item
 LapLDA: Laplacian Linear Discriminant Analysis~\cite{chenjian} presents a least squares formulation for LDA, which intends to preserve both of the global and local structures.
 \item
 LSDA: Locality Sensitive Discriminant Analysis~\cite{caideng} aims to seek a projection which maximizes the margin between data points from different classes at local areas.
\item
 FSSL: This method proposes a framework for joint feature selection and subspace learning \cite{gu2011joint}.
 \item
 MMPP:  Maximum Margin Projection Pursuit \cite{nikitidis2014maximum} aims to find a subspace based on maximum margin principle.
\end{itemize}
\begin{figure*}
\centering
\subfigure[Urban Land Cover]{\includegraphics[width=0.244\linewidth]{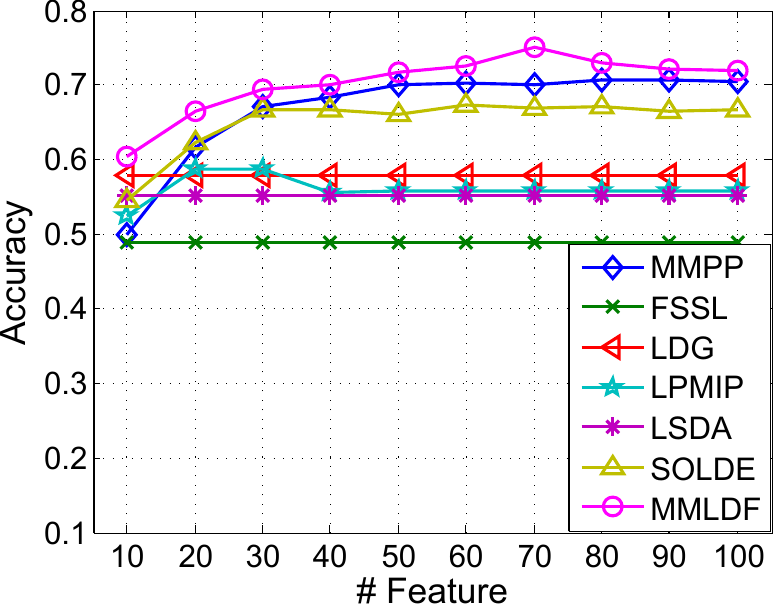}}
\subfigure[CNAE-9]{\includegraphics[width=0.244\linewidth]{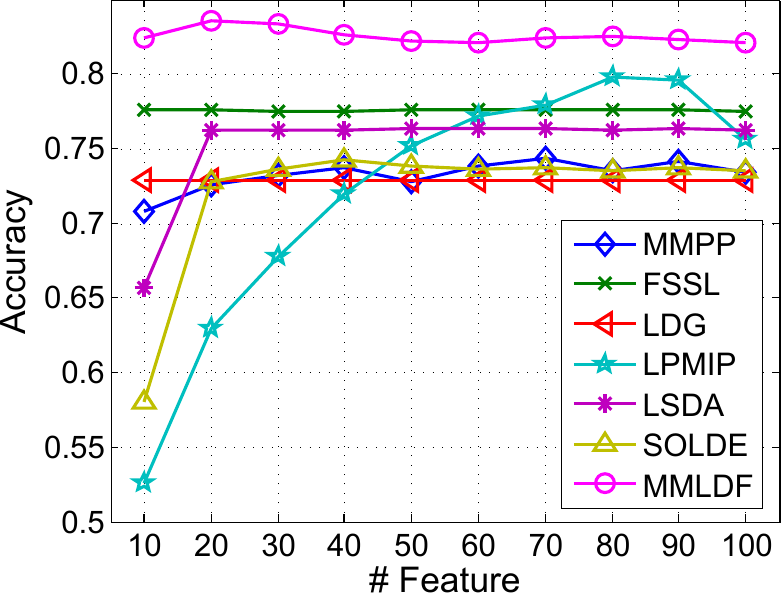}}
\subfigure[DNA]{\includegraphics[width=0.244\linewidth]{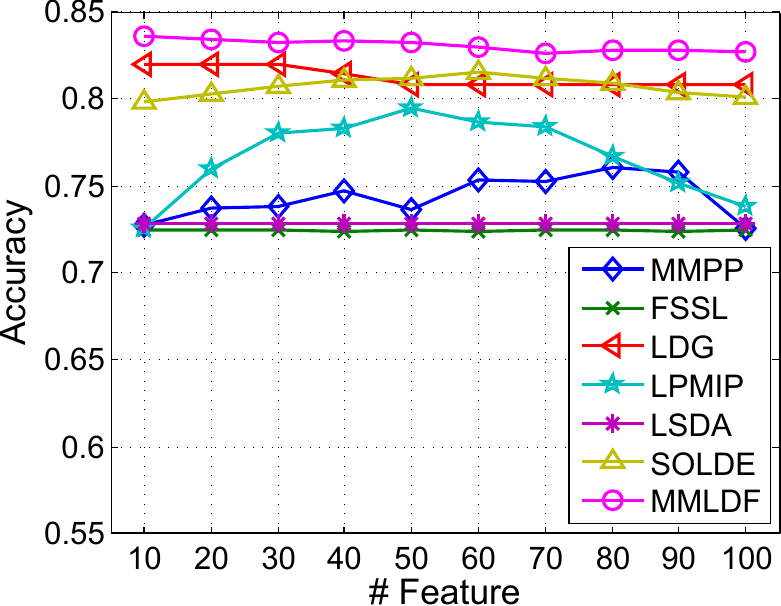}}
\subfigure[Glioma]{\includegraphics[width=0.244\linewidth]{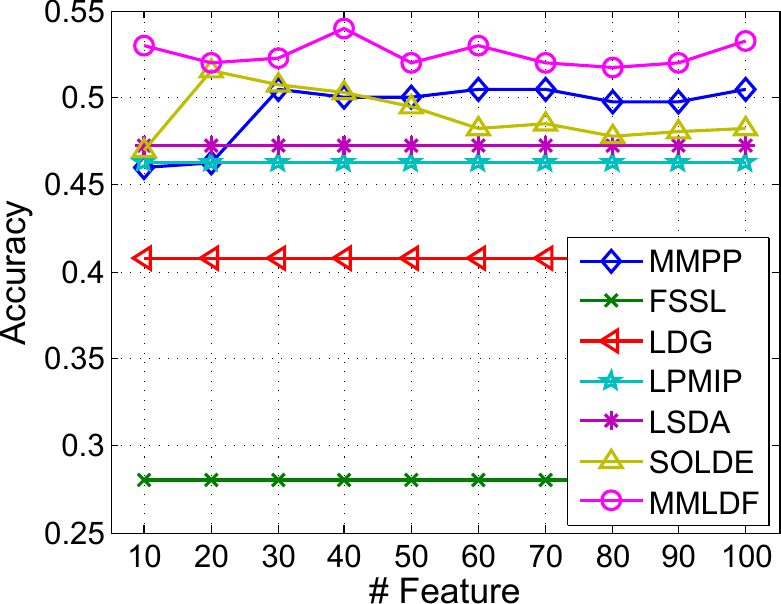}}
\subfigure[LSVT Voice Rehabilitation]{\includegraphics[width=0.244\linewidth]{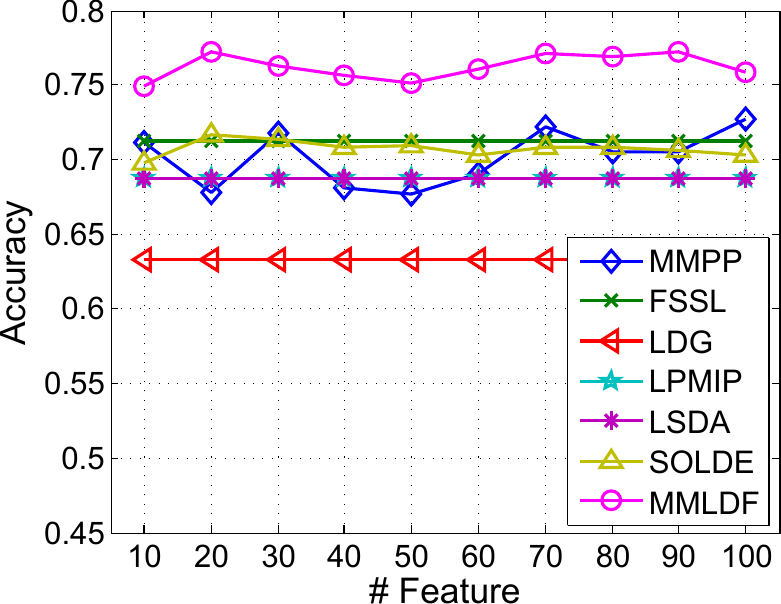}}
\subfigure[Epsilon]{\includegraphics[width=0.244\linewidth]{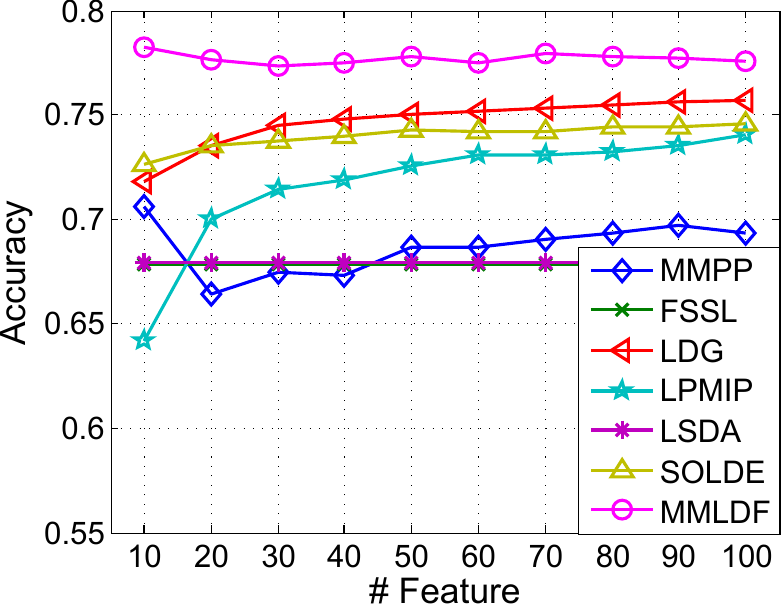}}
\subfigure[Yale Face]{\includegraphics[width=0.244\linewidth]{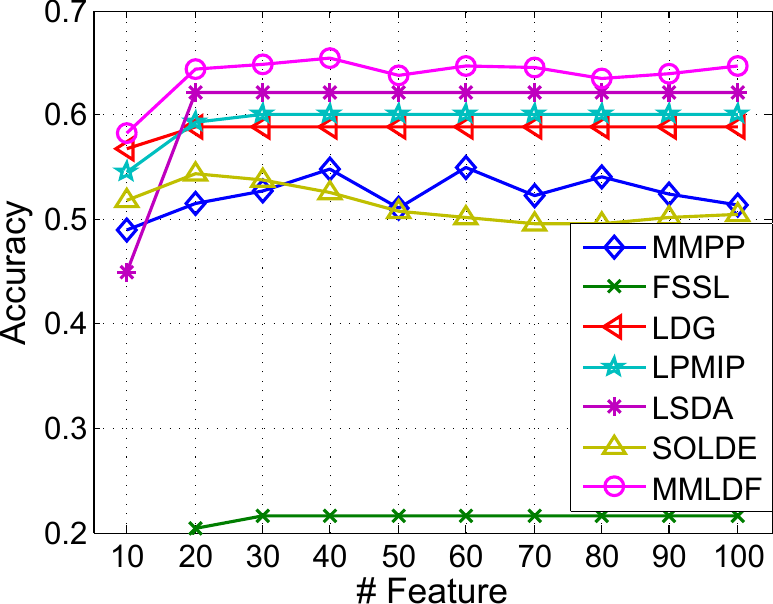}}
\subfigure[15 scene]{\includegraphics[width=0.244\linewidth]{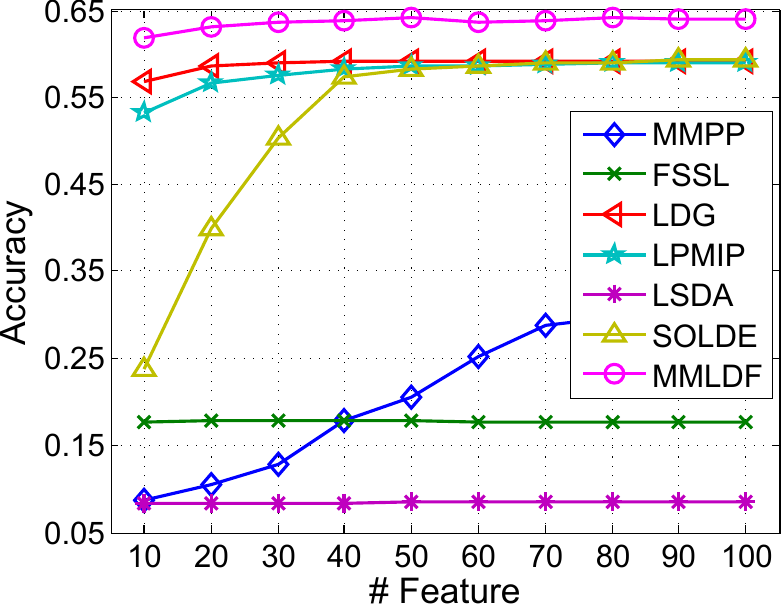}}
\label{feature_accy}
\caption{Classification accuracy of different algorithms vs. the reduced dimensions on all the eight datasets.}
\end{figure*}

To some extent, our method is related to these methods: similar to SOLDE and LSDA, our method aims to preserve the manifold structure in the learned low-dimensional subspace. Moreover, our method tries to preserve global and local information simultaneously, like LPMIP and LapLDA. In our work, we also try to unify feature transformation and feature selection in a framework, inspired by FSSL. Additionally, LDG and MMPP are two latest feature learning methods. LDG tries to preserve the local information during feature transformation, while MMPP is based on subspace learning under maximum margin principle.

In the experiment, we mainly investigate the effectiveness of the learned subspace by leveraging classifier learning. Hence we compare our method with the above feature transformation methods. In order to conduct fair comparisons, we use the same classifier, SVM with linear kernel, to evaluate the subspaces derived by different methods. The classification accuracy is used as the evaluation measure.
In the experiments, we vary the reduced dimensions from 10 to 100 with a stepsize of 10.
In our work, there are four parameters: $C$ and $\lambda$, $\eta$, and $\rho$. The parameters $C$, $\eta$, and $\rho$ are chosen by cross-validation, and the parameter $\lambda$ is always set to $10^{-4}$ (We found when $\lambda=10^{-4}$, the performance was consistently good on all the datasets). For a fair comparison, the parameters in FSSL, LPMIP, LapLDA, LSDA, and MMPP are searched in the same space with that of MMLDF. For all the experiments, we repeat them 10 times, and report the average results.

\subsection{General Performance}
We first evaluate the classification performance of the proposed approach on all the datasets. Table II reports the experimental results of each algorithm with the optimal dimension. The optimal dimensions are listed in the brackets of Table II.
It can be seen that MMLDF consistently outperforms the other seven algorithms on all the eight datasets. Compared with the second best result on each dataset, our method achieves $6.5\%$, $4.8\%$, $2.0\%$, $6.9\%$, $6.5\%$, $3.3\%$, $5.3\%$, and $8.1\%$ relative improvement on the Urban Land Cover, CNAE-9, DNA, Glioma, LSVT Voice Rehabilitation, Epsilon, Yale Face, and 15 scene datasets, respectively. Some baselines obtain considerably poor performance on certain datasets (e.g., FSSL on the Glioma and 15 scene datasets, LSDA on the 15 scene dataset). The reason may be that the generalization ability of these algorithms is limited, making them difficult to be applied to different areas datasets.

We also studied the influence on the performance of different dimensions. Since LapLDA can be only reduced to $\mathcal{K}$ dimensions, where $\mathcal{K}$ denotes the number of the class, we did not compare our method with LapLDA. Fig. 4 shows the results. It can be seen that our method outperforms the other algorithms under all the cases.
\subsection{Analyses on Components' Roles}\label{comp_role}
\begin{figure}[htb]
\centering
\subfigure[LSVT Voice Rehabilitation]{\includegraphics[width=0.49\linewidth]{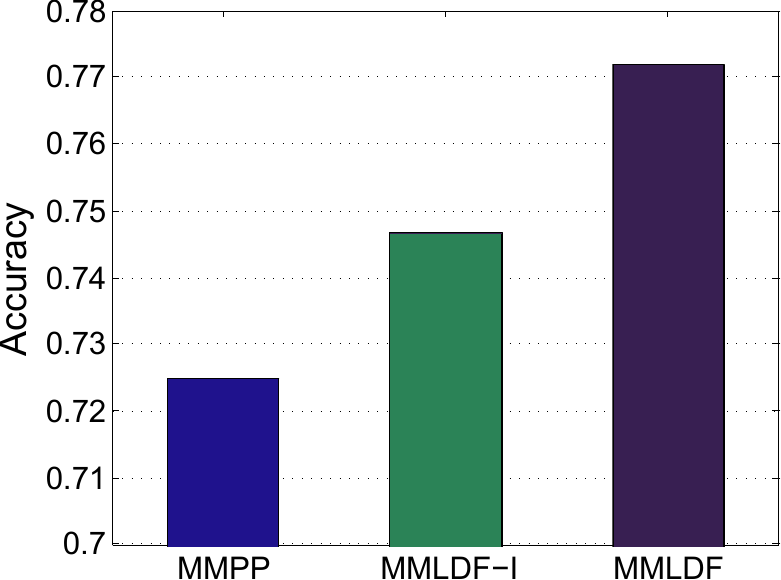}}
\subfigure[Urban Land Cover]{\includegraphics[width=0.49\linewidth]{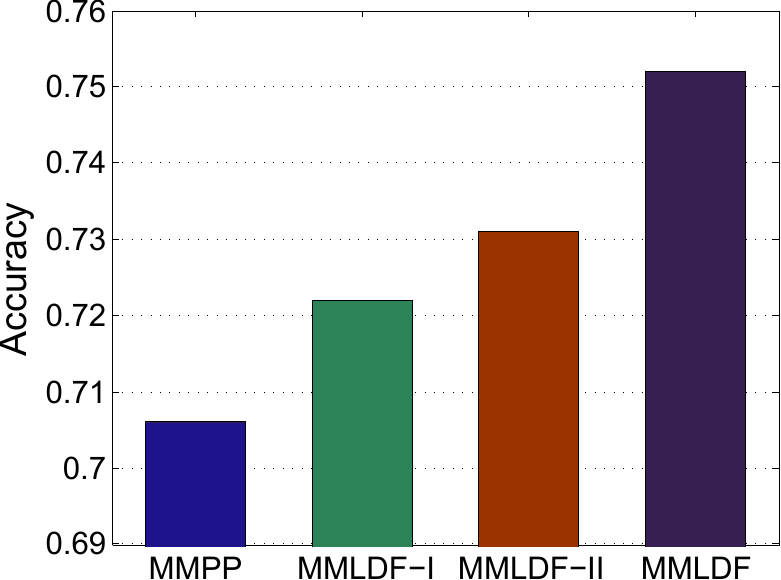}}
\caption{Verify the effectiveness of each component in our algorithm on the LSVT Voice Rehabilitation dataset and the Urban Land Cover dataset.}\label{component}
\end{figure}
We verify the effectiveness of the components in the objective functions (\ref{equ:objg}) and (\ref{total_obj3}), individually. When the parameters $\lambda$, $\eta$, and $\rho$ are set to zeros, MMLDF is reduced to MMPP, thus we use MMPP as the baseline.
We perform the experiments on the binary classification dataset LSVT Voice Rehabilitation and the multi-class classification dataset Urban Land Cover.
The experimental setting is as follows: we first set $\eta$ to zero in (\ref{equ:objg}), and set $\eta$ and $\rho$ to zeros in (\ref{total_obj3}), in order to demonstrate the effectiveness of the module filtering noisy features. We name it MMLDF-I for short.
Then, we set $\rho$ to zero in (\ref{total_obj3}), which indicates that we learn the feature representation without considering the correlations among multiple classes in the case of multi-class classification. We name it MMLDF-II. The dimensions are fixed to 20 and 70 on the LSVT Voice Rehabilitation dataset and the Urban Land Cover respectively, because of better performance of MMLDF under such dimensions on the two datasets based on Table II.

The results are shown in Fig. \ref{component}. We can see that MMLDF-I outperforms MMPP on the datasets, which shows that the row sparseness constraint on the projection matrix is beneficial to learn discriminative feature representation. On the LSVT Voice Rehabilitation dataset, MMLDF achieves better results than MMLDF-I, which means that within-class scatter minimization is good for classification. This point is also verified on the Urban Land Cover dataset, since MMLDF-II is superior to MMLDF-I. On the multi-class dataset, the result of MMLDF is better than that of MMLDF-II, which illustrates that capturing the correlations among multiple categories is helpful for enhancing the discriminative ability of the learned features. MMLDF achieves the best results on both of the two datasets. It shows that the combination of these components is effective for classification.

\begin{table}

\begin{center}
\caption{The robustness (mean+deviation ($\%$)) of $l_{2,1}$ to different noisy levels on the LSVT Voice Rehabilitation dataset.}
\begin{tabular}{|c|c|c|}
\hline
\multirow{2}{*}{ \# Noisy features} & \multicolumn{2}{|c|}{MMLDF}\\
\cline{2-3}
& without $l_{2,1}$-norm & with $l_{2,1}$-norm\\
\hline
50     &71.3$\pm$6.2     &  \textbf{75.2}$\pm$7.3  \\ \hline
100     &70.2$\pm$5.0    &  \textbf{74.5}$\pm$6.8  \\ \hline
150     &69.7$\pm$5.2    &  \textbf{74.4}$\pm$6.8  \\ \hline
200     &68.5$\pm$5.7    &  \textbf{73.0}$\pm$6.1 \\ \hline
\end{tabular}
\end{center}
\vspace{-0.1in}
\end{table}

We further verify the robustness of $l_{2,1}$-norm to different noise levels on the LSVT Voice Rehabilitation dataset.
We generate $N(=50,100,150,200)$-dimensional white Gaussian noise respectively, and concatenate the original features with these noisy features as the new representation of each data. After that, we run MMLDF with $l_{2,1}$-norm and without $l_{2,1}$-norm on the new dataset, respectively. We fix the reduced dimensions to 20, and the experimental results are listed in Table III. We can see the performance of MMLDF with $l_{2,1}$-norm is better than that of MMLDF without $l_{2,1}$-norm, i.e., $l_{2,1}$-norm in MMLDF can indeed alleviate the effects of noisy features on feature transformation.

\subsection {Sensitivity Analysis}
\begin{figure}
\centering
\subfigure[Vary $C$ and reduced dimensions]{\includegraphics[width=0.46\linewidth]{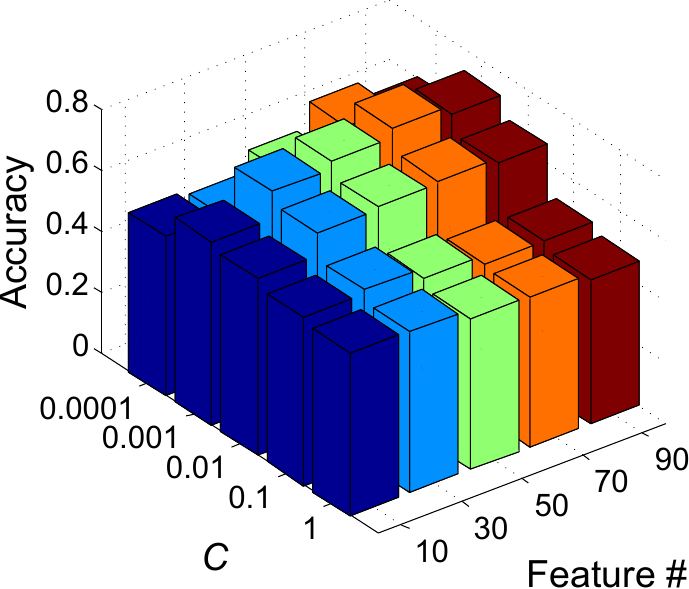}}
\subfigure[Vary $\lambda$ and reduced dimensions]{\includegraphics[width=0.46\linewidth]{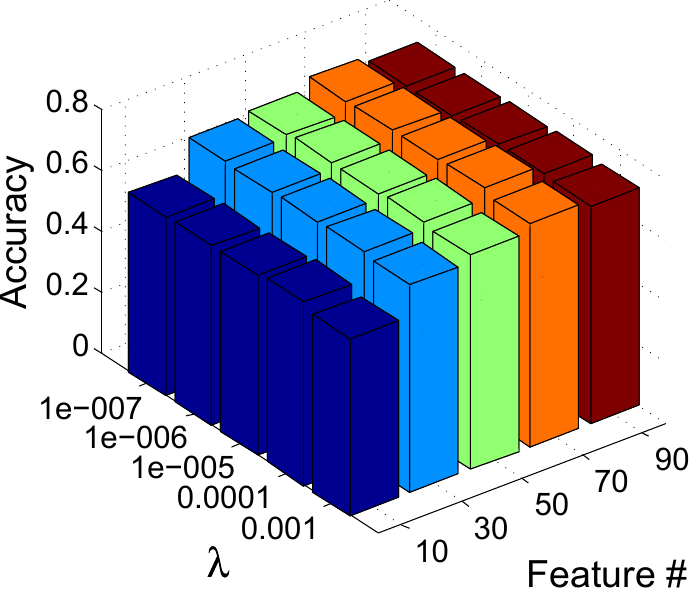}}
\subfigure[Vary $\eta$ and reduced dimensions]{\includegraphics[width=0.46\linewidth]{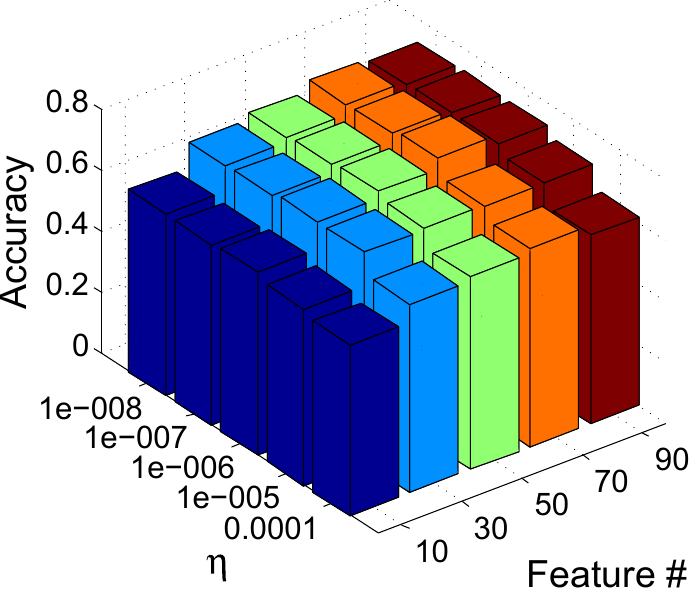}}
\subfigure[Vary $\rho$ and reduced dimensions]{\includegraphics[width=0.46\linewidth]{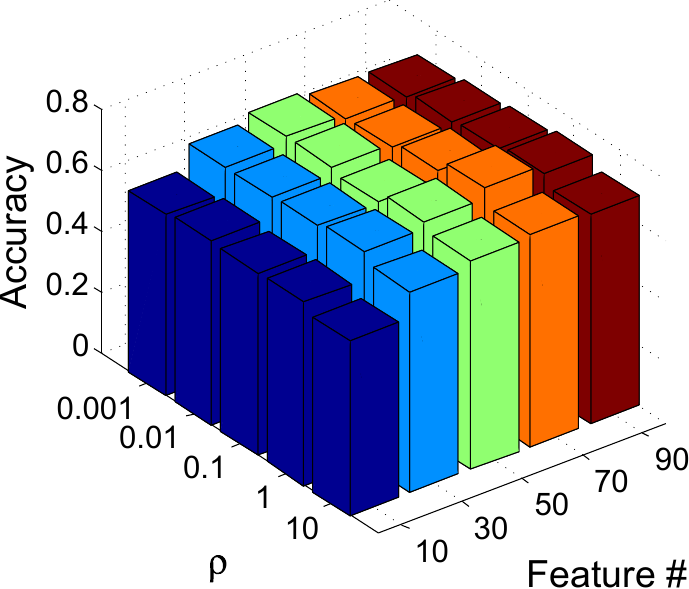}}
\caption{Sensitivity study of the parameters on the Urban Land Cover dataset.}
\label{fig:parameter}
\end{figure}

We also studied the sensitivity of parameters $C$, $\lambda$, $\eta$, and $\rho$ in our algorithm on the Urban Land Cover dataset. Fig. \ref{fig:parameter} shows the results. With the fixed feature dimensions, our method is not sensitive to $\lambda$, $\eta$ and $\rho$ with wide ranges. As for parameter $C$, when we fix the dimensions, the performance is gradually improved as $C$ increases. When $C>10^{-3}$, the performance is gradually degraded as $C$ increases. When $C$ is set to $10^{-3}$, the performance is the best.

\section{Conclusion}
This paper proposed a novel feature transformation method for max-margin based classification. The proposed method aimed to find a low-dimensional feature space to maximize the classification margin of the data and minimize the within class scatter simultaneously. Moreover, we added a regularization term to eliminate noisy or redundant features. Finally, another regularization term was introduced to capture the correlations among multiple categories to help to learn discriminative features. Extensive experiments on publicly available benchmarks demonstrated the effectiveness of the proposed method compared to several related methods.
\bibliographystyle{IEEEtran}
\bibliography{IEEEabrv,egbib}

\end{document}